\documentclass[conference,letterpaper,10pt]{ieeeconf}
\usepackage[left=1.91cm,right=1.91cm,top=1.91cm,bottom=1.91cm]{geometry}

\usepackage{graphicx}
\usepackage{amssymb}
\usepackage{amsmath}
\usepackage{cite}
\usepackage{comment}
\usepackage{color}
\usepackage{url}
\usepackage{alltt}
\usepackage{cleveref}
\usepackage{tikz}
\usepackage{pgfplots}
\usepackage{pgf-umlsd}
\usepackage{cprotect}
\usepackage{multirow}
\usepackage{algorithm}
\usepackage[noend]{algpseudocode}
\usepackage{bigfoot}
\usepackage{paralist}
\usepackage{tabularx}
\usepackage[whole]{bxcjkjatype}
\usepackage{ifthen}
\usepackage{threeparttable}
\usepackage[hang,small,bf]{caption}
\usepackage[subrefformat=parens]{subcaption}
\usepackage[super]{nth}
\usepackage{colortbl}
\usepackage{scalefnt}
\usepackage{threeparttable}

\pgfplotsset{compat=1.14}
\IEEEoverridecommandlockouts

\makeatletter
\newcommand{\figcaption}[1]{\def\@captype{figure}\caption{#1}}
\newcommand{\tblcaption}[1]{\def\@captype{table}\caption{#1}}
\makeatother

\title{\LARGE \bf
Stable Object Placing using Curl and Diff Features\\of Vision-based Tactile Sensors
}

\author{
Kuniyuki Takahashi$^{1}$,
Shimpei Masuda$^{1}$,
Tadahiro Taniguchi$^{2}$
\thanks{$^{1}$K.Takahashi and S. Masuda are with Preferred Networks, Inc.
        {\tt\footnotesize 
        \{{takahashi, masuda\}@preferred.jp}}
        $^{2}$T.Taniguchi is with Ritsumeikan University, College of Information Science and Engineering.
        {\tt\footnotesize 
        taniguchi@em.ci.ritsumei.ac.jp}
        }
}

\begin{document}

\maketitle
\thispagestyle{empty}
\begin{abstract}
Ensuring stable object placement is crucial to prevent objects from toppling over, breaking, or causing spills.
When an object makes initial contact to a surface, and some force is exerted, the moment of rotation caused by the instability of the object's placing can cause the object to rotate in a certain direction (henceforth referred to as \textit{direction of corrective rotation}).
Existing methods often employ a Force/Torque (F/T) sensor to estimate the direction of corrective rotation by detecting the moment of rotation as a torque.
However, its effectiveness may be hampered by sensor noise and the tension of the external wiring of robot cables.
To address these issues, we propose a method for stable object placing using GelSights, vision-based tactile sensors, as an alternative to F/T sensors.
Our method estimates the direction of corrective rotation of objects using the displacement of the black dot pattern on the elastomeric surface of GelSight.
We calculate the \textit{Curl} from vector analysis, indicative of the rotational field magnitude and direction of the displacement of the black dots pattern.
Simultaneously, we calculate the difference (\textit{Diff}) of displacement between the left and right fingers' GelSight's black dots.
Then, the robot can manipulate the objects' pose using \textit{Curl} and \textit{Diff} features, facilitating stable placing.
Across experiments, handling 18 differently characterized objects, our method achieves precise placing accuracy (less than 1-degree error) in nearly 100\% of cases.
\footnote[3]{An accompanying video is available at the following link:\\ \url{https://youtu.be/fQbmCksVHlU}}
\end{abstract}
\begin{figure*}[tb]
	\centering
	\includegraphics[width=1.83\columnwidth]{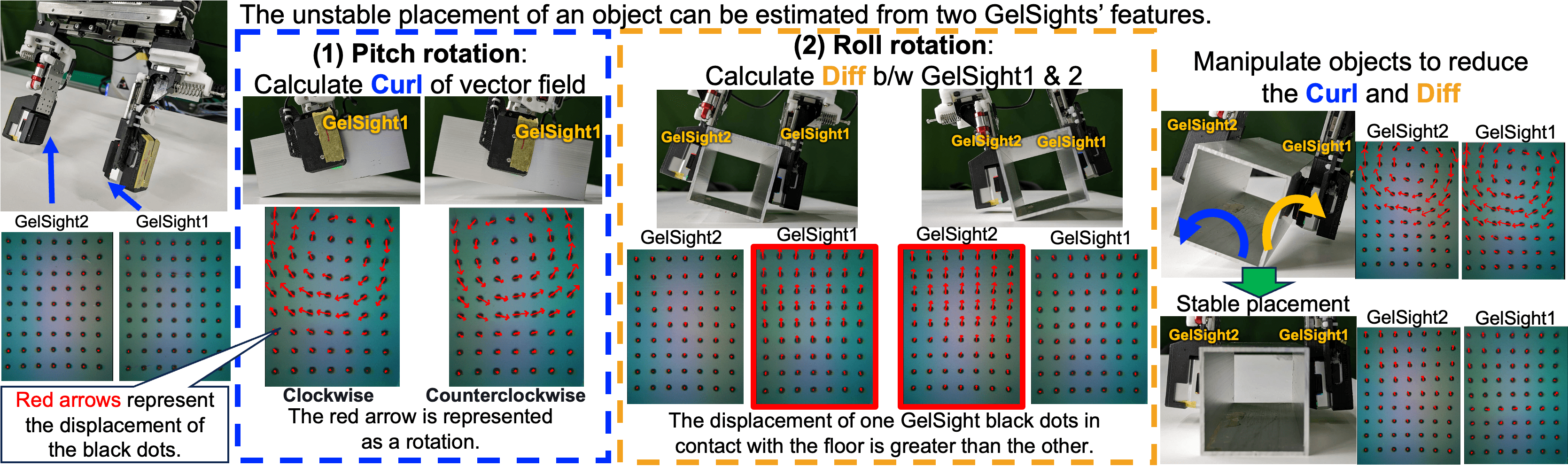}
    \caption{
    Concept of the proposed method:
    The displacement of the black dots on GelSight is analyzed in two patterns corresponding to an object's directions of corrective rotation: roll and pitch.
    By calculating the \textit{Curl} and \textit{Diff} of the displacement of the black dot to estimate the object's direction of corrective rotation and manipulate the object accordingly, the robot ensures its stable placement.
    }
    \label{fig:gelsight_displacement}
    \vspace{-7.0mm}
\end{figure*}
\section{Introduction}
\label{sec:introduction}
Stable placing of grasped objects is important for many reasons, such as 1) preventing objects from toppling over, 2) preventing damage to the objects themselves or their surroundings, and 3) keeping their contents, such as liquids or small particles, from spilling.
Consider settings like chemistry and biology laboratories, for instance, where beakers and flasks are handled routinely.
With the growing demand for laboratory automation, the risk of these objects toppling over with resulting breakage or spillage of their contents has become a significant concern.

We define \emph{stable placing} as a state where a flat-bottomed object's bottom surface is parallel to a table or a non-flat object contacts the table at three or more points.
Conversely, \emph{unstable placing} is defined as the state where a flat-bottom object's bottom surface isn't parallel to the table or a non-flat object contacts the table at less than three points.
Unstable placement can result from many factors, such as inaccurate object pose estimation.

The moment of rotation caused by the instability of the object's placing can cause the object to rotate in a certain direction (henceforth referred to as \textit{direction of corrective rotation}).
The existing method for stable object placing is to use a Force/Torque (F/T) sensor mounted on the robot's wrist by estimating the object's direction of corrective rotation from the torque applied to the grasped object~\cite{raibert1981hybrid}.
However, several issues when using F/T sensors make this estimation uncertain.
1) Tension of external wiring of robot cables: The F/T sensor needs to be calibrated to compensate for the weight of the robot's wrist or gripper.
However, changes in the robot's pose can cause variable tension in the cables, thereby complicating accurate calibration and preventing precise force and torque detection.
2) Sensor noise: The F/T sensor contains noise, which makes the measured sensor values uncertain. 
The F/T sensors attached to the wrist often require a larger rated capacity.
In general, increasing the sensor's rated capacity sacrifices the ability to measure changes in a small range.

To address these issues, typical approaches include mounting a F/T or tactile sensor on the fingertip.
As precision manipulation needs detailed information like fingertip contact area, interest in the use of tactile sensors, particularly vision-based ones like GelSight~\cite{yuan2017gelsight}, are growing~\cite{anzai2020deep, bauza2023tac2pose, dong2021tactile}.
In this context, this study uses GelSight tactile sensors.
Unlike F/T sensors attached to the wrist, tactile sensors are attached to the end-effector's fingertip and are not affected by cable tension from other sensors (such as a camera's USB cable).
Moreover, since detecting small magnitudes of force and torque is required, they are typically designed to minimize the impact of noise.

GelSight uses a camera to measure the deformation of an attached elastomer during contact with a surface.
Some GelSight devices feature black dots on their surface to track their displacement when they make contact with a surface.
The displacement of GelSight's black dots provides significant information, and some previous studies used it for a learning-based approach in peg-insertion task~\cite{zhang2019effective, dong2021tactile}.

Our work proposes utilizing these displacements as an alternative to torque measurements traditionally given by F/T sensors to achieve stable object placing.
Our observations of how tactile sensors respond to various objects during unstable placement suggest that the displacement patterns of GelSight's black dots can be categorized into two types with two directions of corrective rotation: roll and pitch (Fig.~\ref{fig:gelsight_displacement}).
After analyzing each classification, the displacement of GelSight's black dots can be expressed as a one-dimensional feature for each direction of corrective rotation.
The direction of corrective rotation of the object is estimated by calculating the \textit{Curl} from vector analysis for the roll axis, indicative of the rotational field magnitude and direction of the displacement of the black dots pattern, and the difference (\textit{Diff}) of displacement for the pitch axis between the left and right fingers' GelSight's black dots.
These features allow accurate stable object placing on the robot.
The novelty and contribution of this method are as follows:

\begin{itemize}
\item Estimation of the direction of corrective rotation of objects for stable placing using a training-free method employing \textit{Curl} and \textit{Diff} features based on the GelSight's black dots displacement.
To the best of our knowledge, achieving stable placement with proposed \textit{Curl} and \textit{Diff} is a novel and non-trivial achievement.
\item  The use of these features yielded a nearly 100\% success rate of stable placing, achieving a consistently high rate across various objects with less than a 1-degree error. 
Despite the apparent simplicity of our proposed method, it is versatile enough to be applied to many objects.
\end{itemize}
\section{Related Works}
\label{sec:related_works}
\subsection{Object Pose Estimation}
\label{sec:object_pose_estimation}
When considering a placing task, a typical first step is the use of object pose estimation methods.
Object pose estimation is a well-studied problem in computer vision and is important for robotic tasks.
While many approaches are image-based, they often face issues associated with occlusion and coarse accuracy~\cite{sun2022onepose, Li2023vox}.
It is worth noting that there exist methods that reduce uncertainty and increase object pose estimation accuracy by having the grasped object contact other objects~\cite{chavan2018stable, von2020contact, pankert2023learning}.
However, these require prior knowledge of the object's shape, limiting their use for arbitrary objects.

Research utilizing tactile sensors has demonstrated their ability to estimate pose with greater accuracy~\cite{anzai2020deep, lach2023placing}.
In particular, vision-based tactile sensors, such as GelSight, have been used in recent research, and some have achieved success in tasks with milli-order to sub-milli-order accuracy~\cite{anzai2020deep, dong2021tactile, bauza2023tac2pose, pai2023laboratory, ota2023tactile}.
However, these methods only estimate an object's relative or absolute pose and do not consider whether the object can be placed stably on a desk or other surface. 
If there is an error in the estimation result, the object may be unstable and may topple over.
Therefore, the focus of this paper is the stable placing of objects.

\subsection{Object Placing}
\label{sec:object_placing}
Object placing can be approached from both hardware and software perspectives.
A well-known method for the hardware approach is to incorporate passive compliance in the robot's wrist~\cite{whitney1982quasi}.
This mechanical compliance allows the robot to adjust to its environment and achieve stable placing without intricate control manipulations.
However, hardware approaches are often specialized for specific objects, making them less flexible in accommodating a variety of objects.

As for the software approach, conventional methods involve control of applied force using robot joint torque or F/T sensors~\cite{raibert1981hybrid}.
Yet, as discussed in the introduction, these methods often require substantial engineering effort to address problems like cable tension and sensor noise.

Several studies have proposed the use of tactile sensors, either solely or combined with F/T sensors, to estimate and control the force and torque on an object~\cite{ota2023tactile, higuera2023neural,dong2021tactile, zhang2019effective}.
These studies can be broadly divided into learning-based and analytical approaches.
The learning-based approach, such as \cite{ota2023tactile, higuera2023neural}, predicts where and how much external forces are applied to an object and performs object manipulation based on these predictions.
In \cite{dong2021tactile}, a policy for controlling the object pose directly from raw images of GelSight is proposed. 
These studies indicate that tactile sensors can be used as F/T sensors, either directly or indirectly, by learning from raw images of vision-based tactile sensors and flow images showing tactile sensor displacement.
While these approaches can generally adapt to various objects, the prerequisite of extensive training can be time-consuming and may result in out-of-distribution problems if data is insufficient.

In an analytical approach, \cite{zhang2019effective} successfully measured the force and torque applied to a single GelSight sensor by using the Helmholtz-Hodge Decomposition to calculate the relationship between multiple forces and torques applied to different objects and GelSight's black dots displacement.
This method does not require the collection of large-scale data on various objects.
However, to fully capture the potential of these approaches, a conversion to force and torque is attempted, which may necessitate calibration with dedicated equipment.
In our work, we propose an analytical approach that avoids the need for calibration and shows promise in achieving stable object placing across various objects.

\begin{figure}[tb]
    \centering
    \includegraphics[width=0.99\columnwidth]{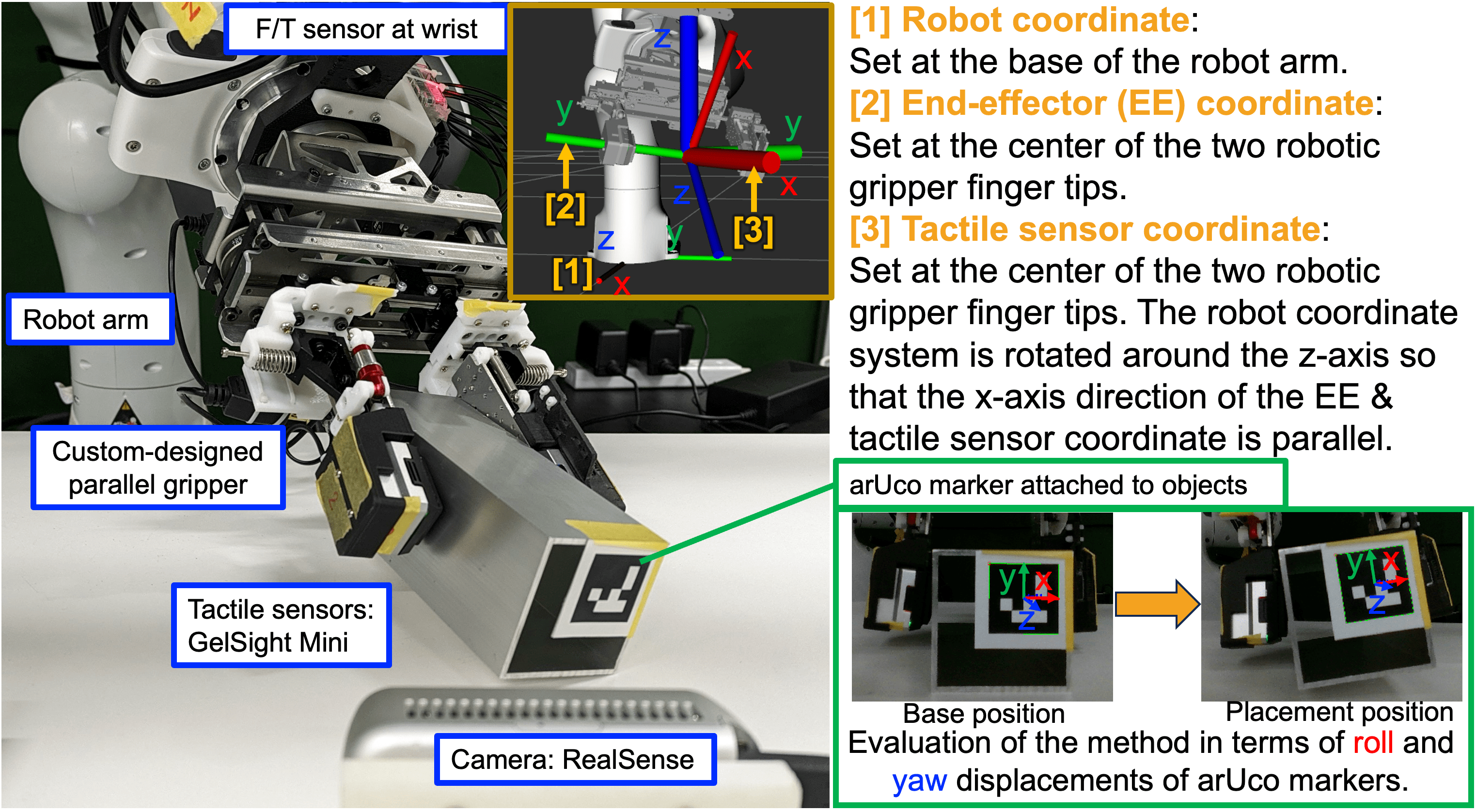}
    \caption{Robot setup and coordinate system used for the experiment}
    \label{fig:robot_setup}
    \vspace{1.0mm}
\end{figure}
\section{Preliminaries}
\label{sec:Preliminaries}
As Preliminaries, this section describes the estimation of the direction of corrective rotation of an object when the object is placed unstably by pressing the object against a desk.
Then, the robot control method for stable placing based on sensor information is described.

Three coordinate systems used in this study are described below (Fig.~\ref{fig:robot_setup}).
The first one is the \textbf{robot coordinate} system, which is set at the base of the robot arm.
The second is the \textbf{end-effector (EE) coordinate} system, which is set at the center of the two robot gripper's fingertips.
The robot manipulates objects by applying force and torque to them for stable placing.
The point where the robot arm generates force is the origin of the EE coordinate system.
In this paper, `the point where the robot arm generates force' is referred to as the \textit{force application point}.
The third is the \textbf{tactile sensor coordinate} system, in which the robot coordinate system is rotated around the z-axis so that the x-axis direction of the tactile sensor coordinate system is parallel to the x-axis direction of the EE coordinate system.
\begin{figure}[tb]
    \centering
    \includegraphics[width=0.90\columnwidth]{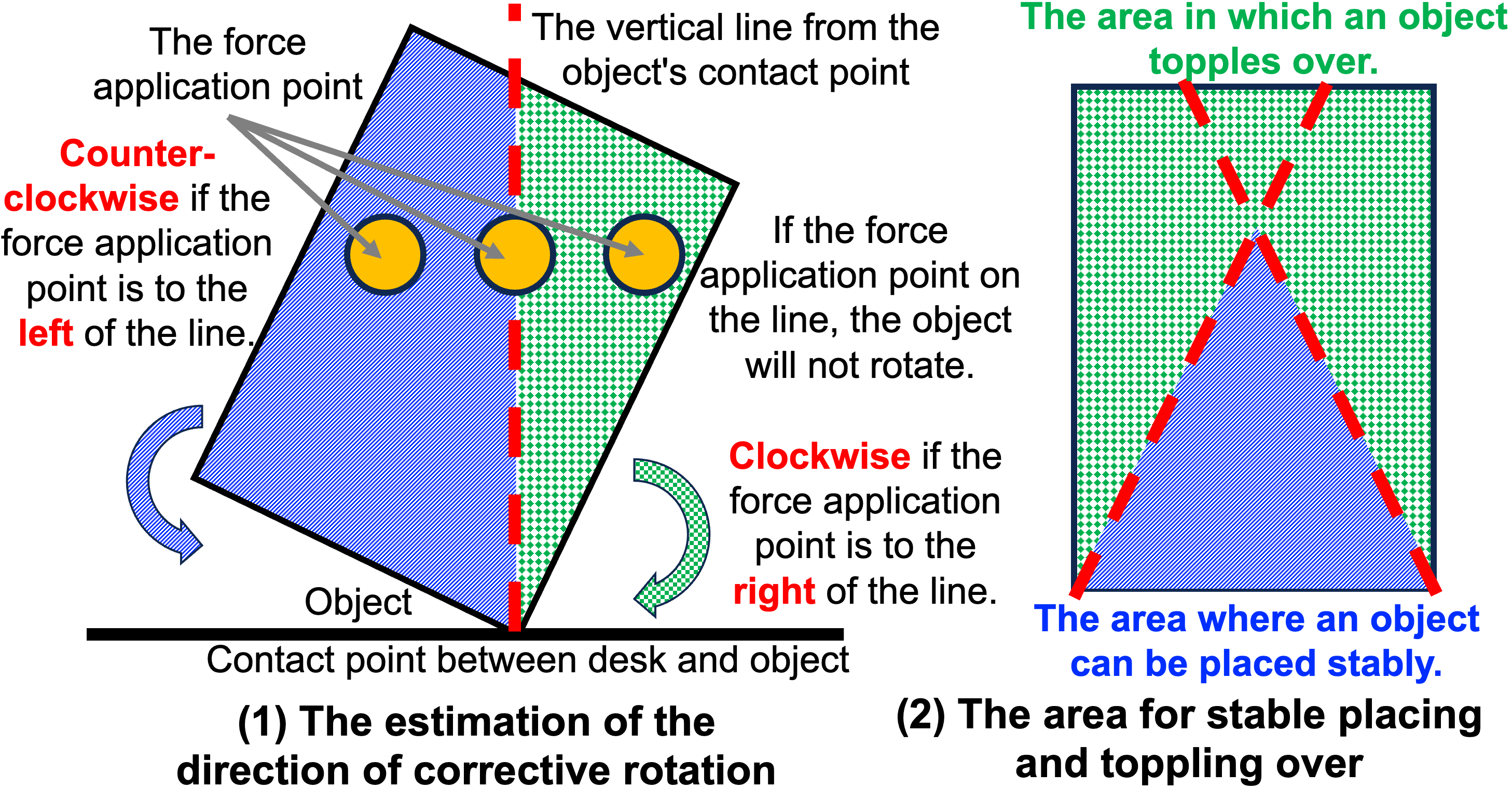}
    \caption{The estimation of the direction of corrective rotation of an object when the object is pressed against a desk.}
    \label{fig:rotation_direction}
\end{figure}
\subsection{Estimation of Direction of Corrective Rotation}
\label{sec:Estimation of Direction of Corrective Rotation}
The idea behind the estimation of the direction of corrective rotation can be explained by considering the moment around the point of contact between the object and the environment~\cite{shimizu2002spatial}.
The direction of corrective rotation is determined based on whether the \textit{force application point} is to the right or left of the vertical line from the object's contact point (Fig~\ref{fig:rotation_direction}~(1)).
The direction of corrective rotation is clockwise when the \textit{force application point} is on the right side of the vertical line and counterclockwise when the \textit{force application point} is on the left.
If the \textit{force application point} is on the vertical line, the object does not rotate.

The sign of the torque at the \textit{force application point} indicates the direction of corrective rotation.
The torque is positive, indicating clockwise rotation, if the \textit{force application point} is to the right side of the vertical line; negative, for counterclockwise rotation, if on the left.
We note that depending on the convention used, positive torque can indicate either clockwise or counterclockwise rotation but should stay consistent across future experiments.
If the torque is zero, the \textit{force application point} is on the line, indicating that the object does not rotate.
Based on the sign of this torque detected by sensors, the robot is controlled to perform stable placing of the object.
This control method of the robot is described in \Cref{sec:Velocity Control of End-effector}.

Sensors measure these torques to estimate the direction of corrective rotation.
While the method using the F/T sensor (referred to in this paper as the \textbf{F/T sensor-based method, which is the baseline} for this paper) can estimate the direction of corrective rotation, it is prone to inaccuracies primarily due to the cable tension and sensor noise.
Especially near the vertical line, misrecognition of the direction of corrective rotation may occur as the measured sign of torque is likely to invert.
Our proposed method uses tactile sensors to overcome these issues, providing a more accurate estimation of the direction of corrective rotation.
Although tactile sensors measure different physical quantities than torque sensors, their measurements can still be interpreted as positive or negative with respect to the vertical line, providing a consistent approach to the estimation of the direction of corrective rotation.

\subsection{Robot Control for Stable Placing}
\label{sec:Velocity Control of End-effector}
The previous section described that the direction of corrective rotation of the object can be estimated based on the sign of the torque.
This section describes how to control the robot for stable placing based on the torque.

If the torque measured at the force application point is not zero, the object, which is in an unstable state, can be rotated in the direction of corrective rotation.
Stable placing can be performed by controlling the object by applying torque to the \textit{force application point} so that the torque becomes 0.
We make assumptions for our model, which are detailed below:
As shown in Fig.~\ref{fig:rotation_direction}~(2), the force application point is in the area where the object can be placed stably.
We note that the area where this stable placement is feasible is not constant but varies depending on the position of the vertical line from the object's contact point at the time the object is unstably placed.
Under this assumption, manipulating the object so that the detected torque is zero results in a stable placing.

To realize stable placing, the robot arm is controlled by admittance control so that force $F_{target} = (F_{target-x}, F_{target-y}, F_{target-z})=(0, 0, F_{tz})$ and torque $\tau_{target} = (\tau_{target-x}, \tau_{target-y}, \tau_{target-z})=(0, 0, 0)$ for the \textit{force application point} exerts the force and torque in the robot coordinate system.
The combination of force and torque is called a wrench.
Let $w_{target} = (F_{target}, \tau_{target})$ be the target wrench and $w_{current} = (F, \tau)$ be the current wrench.
For this target wrench, when the object is pushed against the desk, the reaction force from the desk is $F_{tz}$, and the other forces and torques are 0.
In this experiment, $F_{tz} =  5\mathrm{\,N}$.

The end-effector velocity of the robot arm, $v_{ee}$ from the EE coordinate system, is calculated as follows to achieve the target wrench $w_{target}$:
\begin{equation}
    v_{ee} = k ( w_{current}-w_{target})
    \label{eq:velocity_control}
\end{equation}
where $k$ is a gain.
In this paper, the same value for this gain is used for all objects.

In the case of the F/T sensor-based approach, the current wrench $w_{current}$ uses an F/T sensor mounted on the wrist.
Let's denote the force measured by the F/T sensor on the wrist as $F'$ and the torque as $\tau'$.
The force $F$ and torque $\tau = (\tau_{x}, \tau_{y}, \tau_{z})$ at the \textit{force application point} can be calculated as follows:
\begin{align}
    F &= R \cdot F'  \nonumber\\
    \tau &= R \cdot \tau' + d \times (R \cdot F')
    \label{eq:force_torque}
\end{align}
Here, $R$ is a 3x3 rotation matrix, and $d$ is a 3x1 displacement vector from the F/T sensor coordinate to the EE coordinate in the F/T sensor coordinate. 

In our method, we use the \textit{Curl} and \textit{Diff} features values calculated by GelSight, instead of the $\tau_{x}$ and $\tau_{y}$ values for roll and pitch that are calculated from the F/T sensor.

\section{Method: Tactile-based Placing}
\label{sec:tactile-based manipulation}
The previous section described that the direction of corrective rotation can be estimated based on the sign of the torque.
This section describes how to use tactile sensors as an alternative to F/T sensors to estimate the direction of corrective rotation.

Our observations of how tactile sensors respond to various objects during unstable placement suggest that the displacement patterns of GelSight's black dots can be categorized into two types with two directions of corrective rotation: roll of the tactile sensor coordinate system and pitch of the EE coordinate system (Fig.~\ref{fig:gelsight_displacement}).
The following section describes how to calculate the rotations for pitch and roll based on the displacement of the black dots on the GelSight.
\begin{figure}[tb]
    \centering
    \includegraphics[width=0.90\columnwidth]{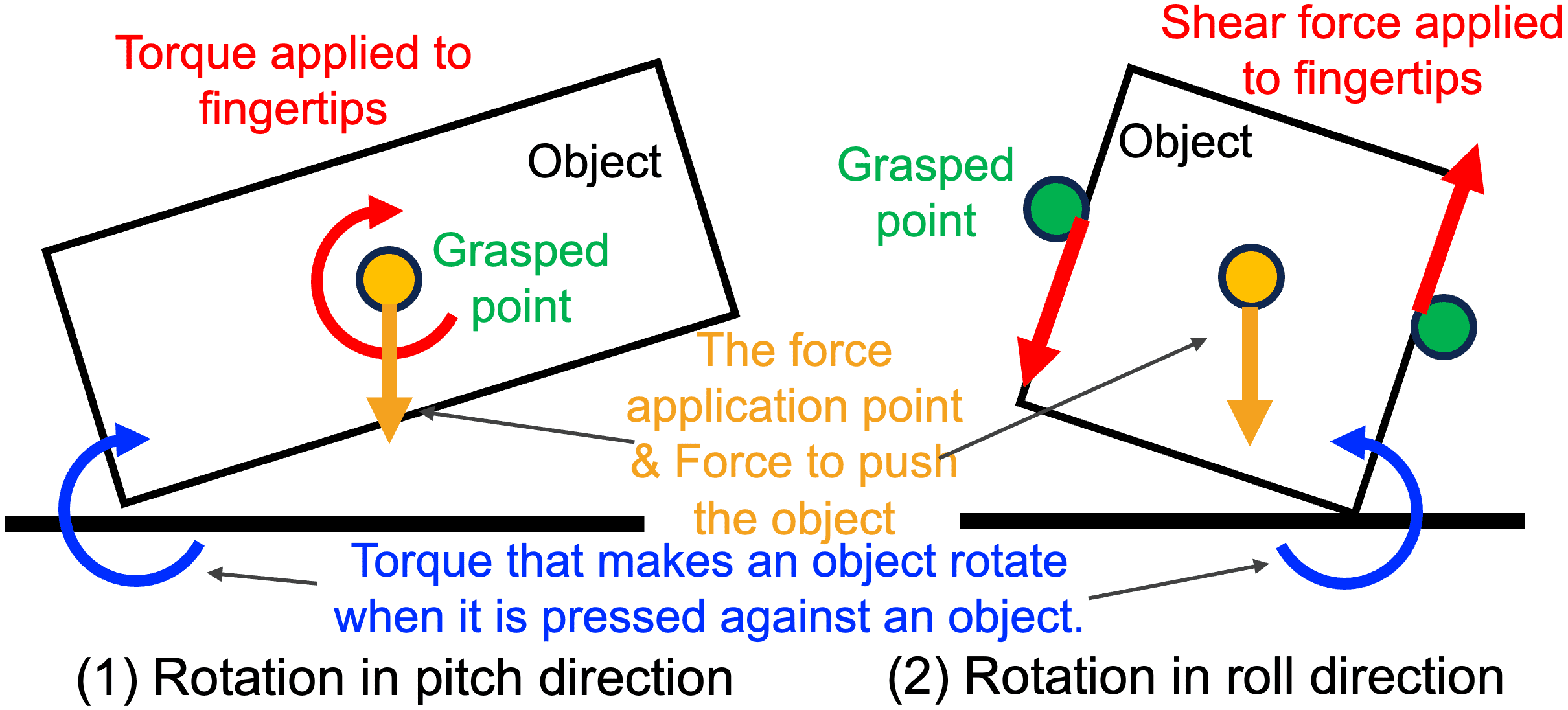}
    \caption{Schematic of torque applied to fingertip}
    \label{fig:torque}
\end{figure}
\subsection{Calculation of Rotation in Pitch Direction by Curl}
\label{sec:Calculation of Rotation in Pitch Direction}
We describe a method where the \textit{Curl} features calculated from the displacement of the black point of the tactile sensor, used as an alternative to $\tau_{y}$ (pitch rotation at the force application point) calculated from the F/T sensor by \cref{eq:force_torque}.

When the object is pressed against the desk, a torque is generated that attempts to rotate the object (Fig.~\ref{fig:torque}~(1)).
This torque is transmitted to the GelSight and results in a shift in the position of the black dots on the GelSight, which represents the applied torque.
This torque at the contact point on GelSight can be observed as the displacement of the black dots on GelSight (Fig.~\ref{fig:gelsight_displacement}~(1) Pitch rotation).
From the red arrows in Fig.~\ref{fig:gelsight_displacement}~(1), when the left side of the object contacts the desk (with the gripper considered as the center), we observe a displacement of the arrows representing the black dots on the GelSight moving in a specific rotational direction.
If the right side of the object is in contact instead, the arrows exhibit the opposite rotational movement.
Specifically, in the former case, the arrows move in a clockwise direction, whereas in the latter, they move in a counterclockwise direction.

When the displacement of these arrows, which represent the movement of the black dots on the GelSight, becomes zero, the object is in a stable placing state with respect to the pitch rotation.
To achieve a stable placement of the object by minimizing the torque, the robot should rotate around the \textit{force application point} in the direction indicated by these arrows, which in turn reduces the displacement of the black dots.
We note that the contact point is the fingertip, and the \textit{force application point} is the middle of the contact point between the two fingers.
Thus, this rotation center is considered the origin of the EE coordinate system.

From the arrows, the direction of rotation and its degree of rotation are calculated by \textit{Curl} of vector analysis.
The Curl is indicative of the rotational field magnitude and direction of the displacement of the black dots pattern.
The displacements of the black dots on the GelSight give a vector field $A_{tactile} = (A_{x}, A_{z})$.
In the EE coordinate system, the Curl of this field can be computed as follows:
\begin{equation}
    \text{Curl} \: A_{tactile} = \partial A_{z}/\partial x - \partial A_{x}/\partial z
    \label{eq:curl}
\end{equation}
We use the mean of the calculated \textit{Curl} of all black dots as an alternative to $\tau_{y}$ calculated from the F/T sensor by \cref{eq:force_torque}, the torque at the \textit{force application point}.
The goal is to control the end-effector speed by \cref{eq:velocity_control} of the robot so that this value becomes $0$.

\subsection{Calculation of Rotation in Roll Direction by Diff}
\label{sec:Calculation of Rotation in Roll Direction}
We describe a method where the \textit{Diff} features calculated from the displacement for the pitch axis between the left and right fingers' GelSight's black dots, used as an alternative to $\tau_{x}$ (roll rotation at the force application point) calculated from the F/T sensor by \cref{eq:force_torque}.

When the object is pressed against the desk, a torque is generated that attempts to rotate the object (Fig.~\ref{fig:torque}~(2)).
This torque is transmitted to the GelSight, and is exerted as shear forces in opposite directions on the right and left GelSight.
This makes a difference in the displacement of the black dots in GelSights between the left and right fingers.
The shear force arising at the contact point on GelSights can be observed by the displacement of the black dots on the GelSights (Fig.~\ref{fig:gelsight_displacement}~(2) Roll rotation).
Additionally, there are also forces generated by the action of pushing the object against the table that is exerted on the GelSights, whose magnitudes may change the overall direction of displacement.
Fig.~\ref{fig:gelsight_displacement}~(2) demonstrates the displacement of the black dot of GelSight when the tilted object makes contact with the desk.
On the side where the object is in contact, the displacement of the black dot is represented as an arrow pointing vertically upward (in the z-axis direction of the tactile sensor coordinate system).
This displacement is noticeably larger compared to the side where the object is not making contact.

When there is no longer any imbalance between the left and right of the displacement of the black dot of GelSight, the object is in a state of stable placing with respect to the direction of rotation of the roll.
In order to reduce the imbalance of the left and right of the displacement of the black dot of GelSight, the robot should rotate around the \textit{force application point} with respect to the direction of the object's rotation.
We note that the contact point is the fingertip, and the \textit{force application point} is the middle of the contact point between the two fingers.
Thus, this rotation center is considered the origin of the EE coordinate system.

In this study, we use two GelSight sensors, GelSight1 and GelSight2.
Considering the black dots' displacement in the z-axis direction of the tactile sensor coordinate system (normal facing upwards to the desk) helps calculate the GelSights' displacement affecting the roll axis.
This makes it possible to be irrespective of the pose of the robot and GelSights.
We compute the average displacement values of the black dots on each of the two GelSights along the z-axis of the tactile sensor coordinate system.
We denote these values as $\bar{A}_{z1}$ and $\bar{A}_{z2}$ respectively.
We define a metric \textit{Diff} as the difference between $\bar{A}_{z1}$ and $\bar{A}_{z2}$:
\begin{equation}
    Diff = \bar{A}_{z1} - \bar{A}_{z2}
    \label{eq:diff}
\end{equation}
\textit{Diff} serves as the torque at the \textit{force application point} in the tactile sensor coordinate system.
The goal is to control the robot's end-effector speed to reduce \textit{Diff} to zero.
We note that the end-effector speed calculated from \textit{Diff} corresponds to the speed measured from the tactile sensor coordinate system.
However, as $\tau_{y}$ in \cref{eq:force_torque} is calculated in the EE coordinate system, we must convert the value of \textit{Diff} from the tactile sensor to the EE coordinate system.
This can be achieved using standard coordinate transformation methods.
\section{Experimental Setup}
\label{sec:experimental setup}
\subsection{Robot Setup}
\label{sec:Robot Setup}
Our robotic system, as shown in Fig.~\ref{fig:robot_setup}, features a 7-DoF Franka Emika Panda Arm.
The end-effector command to the robot is updated at $200 \mathrm{\,Hz}$.
This robotic arm is equipped with a custom-designed parallel gripper.
The gripper features vision-based tactile sensors, known as GelSight Mini~\cite{gelsight2024}.
The sampling rate of GelSights is $10 \mathrm{\,Hz}$.
A Leptrino force/torque sensor (FFS055YA501U6) is mounted between the robot arm and the gripper.
The rated capacity of force of F/T sensor is $F_{xyz}= \pm 500 \mathrm{\,N}$, torque is $T{xyz}= \pm 4 \mathrm{\,Nm}$, resolution is $\pm1/2000$, and sampling rate is $200 \mathrm{\,Hz}$.
An Intel RealSense Depth Camera D435i is used to capture the pose of the arUco marker~\cite{garrido2014automatic}.
The system runs on a PC equipped with 32\,GB RAM and an Intel Core i7-8700 CPU, running Ubuntu 20.04.6 LTS with ROS Noetic.
\begin{figure}[t]
    \centering
    \includegraphics[width=0.85\columnwidth]{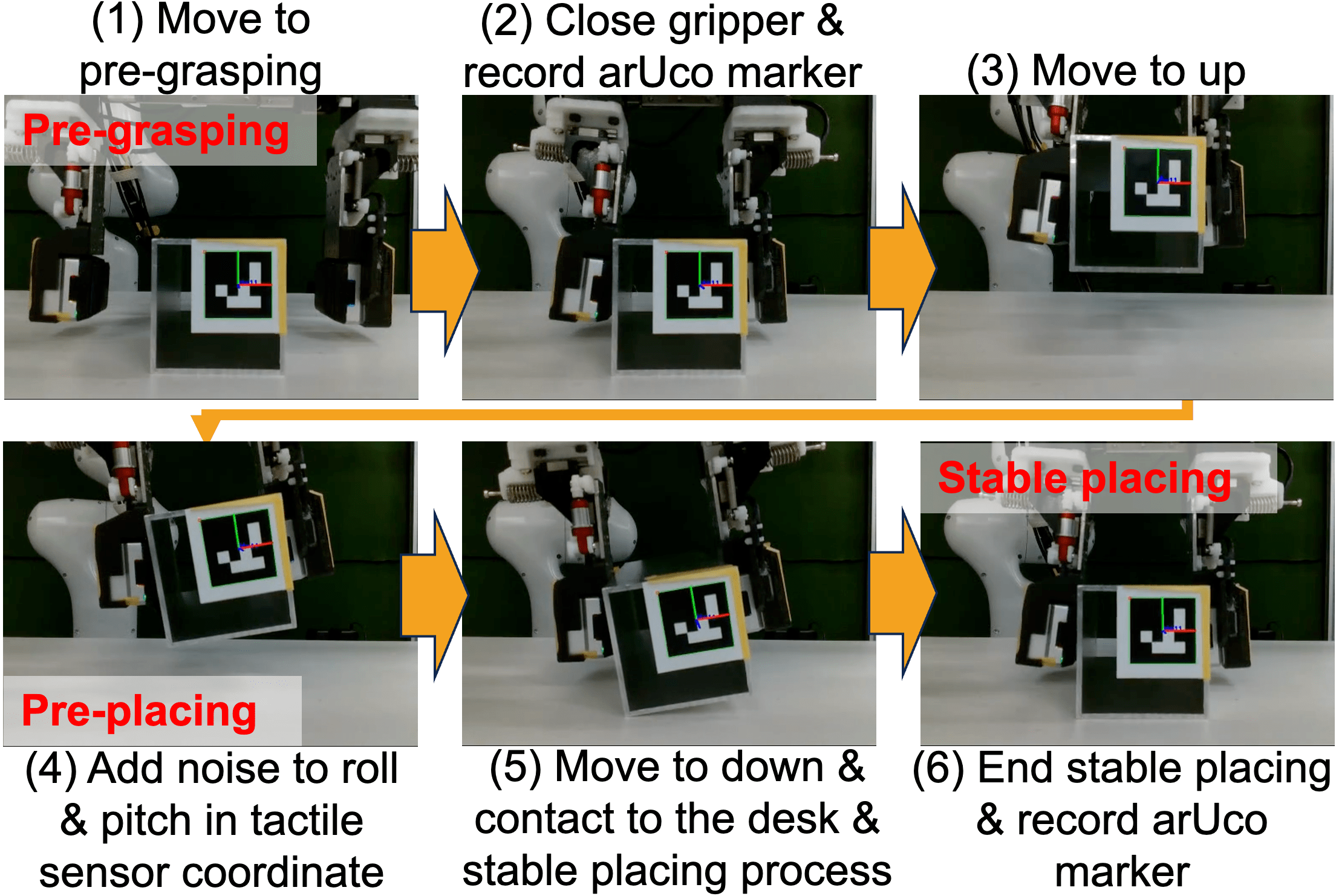}
    \caption{Experiment process for object placing}
    \label{fig:experiment_process}
\end{figure}
\begin{figure*}[t]
    \centering
    \includegraphics[width=1.95\columnwidth]{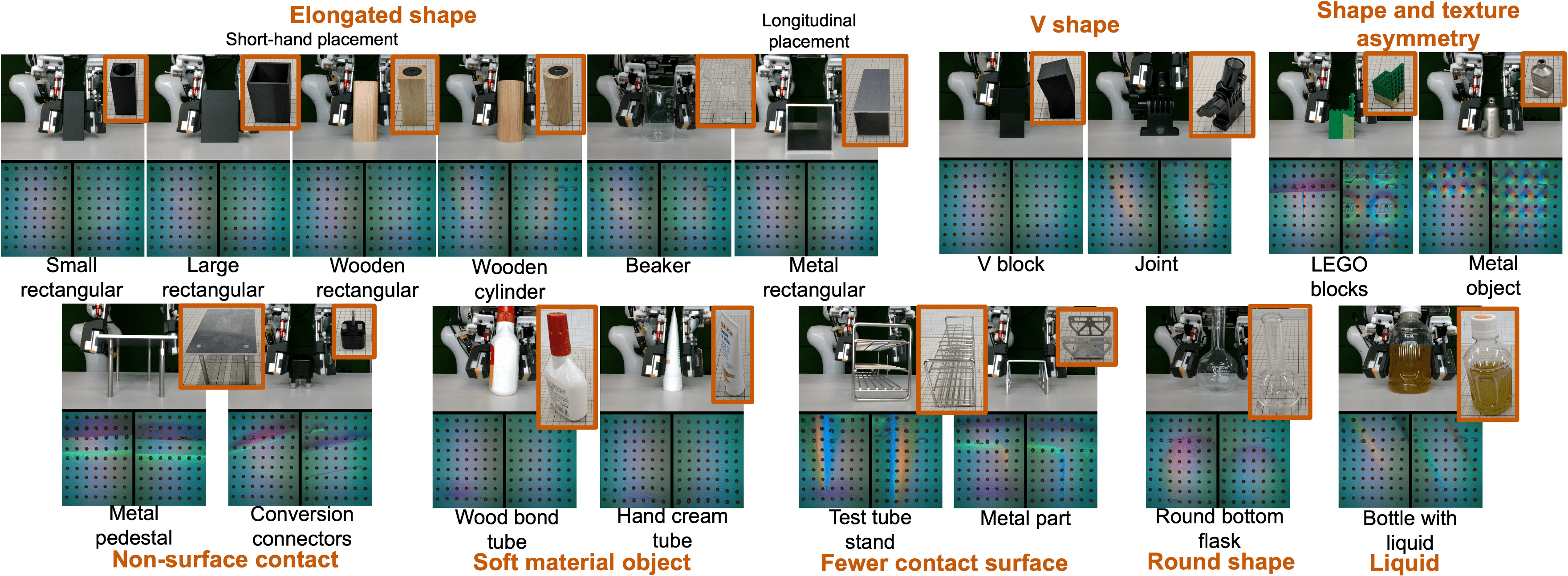}
    \caption{Target objects used in the experiment and GelSight images while grasping the target objects}
    \label{fig:target_objects}
    \vspace{-7.0mm}
\end{figure*}
\subsection{Task Setup and Process of Object Placing}
\label{sec:Process of object placing}
We conduct experiments where a robot grasps objects and places them on a desk to evaluate our proposed system. 
To aid evaluation, an arUco marker is placed on the object, and the object's pose is measured when the object is placed on the desk.
The displacement is then calculated by comparing the object's pose initially in its manually assigned stable state on the table with the final pose achieved through the robot's stable placing process.
The detailed steps of the experimental procedure are graphically illustrated in Fig.~\ref{fig:experiment_process}.

This experiment begins by manually placing an object with an attached arUco marker of size $25~\mathrm{\,mm}$ onto a desk.
We note that the experiment is repeated 10 times for each of the proposed and baseline methods but for the same object placed in the same location.
The robot moves to a predetermined pre-grasping position (Fig.~\ref{fig:experiment_process}~(1)).
The pre-grasping positions are specific to each object.

The robot grasps the object, and then, the camera (RealSense) in front of the robot (Fig.~\ref{fig:robot_setup}), with a sampling rate of $30 \mathrm{\,Hz}$, is used to capture the pose of the arUco marker (Fig.~\ref{fig:experiment_process}~(2)).
The initially placed object is defined as being in a \textbf{stably placed state} in this experiment.
To ensure accuracy and reduce potential noise during marker recognition, we use the average pose from the marker across 90 static images captured by the camera over three seconds.

After capturing the arUco marker, the robot lifts it off the ground by moving along the z-axis, and tilts it (Fig.~\ref{fig:experiment_process}~(3)-(4)).
To evaluate the stable placing, the grasped object is tilted (either fixed or randomly sampled) in this pre-placing process.
The way the tilt is specified is different for each experiment and will be described in the \cref{sec:results}.
In the tilted state, the value read from the F/T sensor is recorded as an offset.
During control procedures, we will use the value derived from the F/T sensor after removing this offset.
The displacement of the black dots of the tactile sensor is calculated from the position of the black dots of the tactile sensor at this stage.

The robot then moves under velocity control at $0.005~\mathrm{\,m/s}$ in the -z-axis direction in the robot coordinate system (Fig.~\ref{fig:experiment_process}~(5)). 
When the force value of the F/T sensor along the z-axis of the robot coordinate system exceeds $1 \mathrm{\,N}$, the process of stable placing begins using both the proposed and baseline methods (Fig.~\ref{fig:experiment_process}~(5)).

After sufficient time has passed to ensure the completion of stable placing, the placing process is stopped, and the pose of the arUco marker is captured again (Fig.~\ref{fig:experiment_process}~(6)).
As with the acquisition of the arUco marker taken in the beginning, the mean value of the 6d pose of the arUco marker was calculated from 90 static images captured over three seconds.
Under the control of stable placing, the displacements in the roll and yaw axes of the arUco marker are evaluated.
This evaluation is done by comparing the marker's pose before and after the stable placing procedure is completed (Fig.~\ref{fig:robot_setup}).

Finally, the robot releases the object and returns to the home position.
We performed this experimental process ten times for each object and calculated the mean and variance of the displacements of the arUco marker at that time.

\subsection{Target Objects}
\label{sec:Target Objects}
For the evaluation, we chose 18 objects considering various factors such as size (lengths of the short and long axes and thickness), softness, surface friction, texture, left-right asymmetrical shape and texture, and whether the object had a flat contact surface or not (Fig.~\ref{fig:target_objects}).
\section{Experiment Results}
\label{sec:results}
This evaluation focused on two key aspects:
1) Evaluation of stable placing, considering changes in the object's grasp positions and the size of its \textit{support polygon} - the area defined by the contact points between the object and the surface upon which it rests.
2) Comparison of the proposed method, \textbf{Tactile-based method}, with baseline,  \textbf{F/T sensor-based method} across a variety of objects.
\begin{table*}[t]
    \centering
    \caption{
            The accuracy of stable placing when the grasping position and the size of the support polygon of the object are different.
            }
    \label{tab:ablation_study}
    \begingroup
    \scalefont{0.90}
        \begin{tabular}{c|c||ccccc|ccccc}
        \hline
            \multicolumn{1}{c|}{\multirow{2}{*}{Object}} &
            \multicolumn{1}{c||}{\multirow{2}{*}{\shortstack{Grasp \\ height}}} &
            \multicolumn{5}{c}{Tactile-based (Ours)} &
            \multicolumn{5}{|c}{F/T-based (Baseline)} \\
            \cline{3-12} 
            \multicolumn{1}{c|}{} & \multicolumn{1}{c||}{}
                             & \multirow{1}{*}{Roll [deg]}
                             & \multirow{1}{*}{Pitch [deg]}
                             & \multirow{1}{*}{All [deg]}
                             & \multirow{1}{*}{$< 1$ deg}
                             & \multirow{1}{*}{$< 2$ deg}
                             & \multirow{1}{*}{Roll [deg]}
                             & \multirow{1}{*}{Pitch [deg]}
                             & \multirow{1}{*}{All [deg]}
                             & \multirow{1}{*}{$< 1$ deg}
                             & \multirow{1}{*}{$< 2$ deg} \\

            \hline \hline
\multicolumn{1}{c|}{\multirow{2}{*}{Large  rectangular}}
& $1.0~\mathrm{\,cm}$ & $0.4\pm0.2$ & $0.2\pm0.2$ & $0.3\pm0.2$ & $\bf{10/10}$ & $\bf{10/10}$ & $0.4\pm0.2$ & $0.2\pm0.2$ & $0.3\pm0.2$ & $\bf{10/10}$ & $\bf{10/10}$\\
& $2.5~\mathrm{\,cm}$ & $0.4\pm0.3$ & $0.2\pm0.1$ & $0.3\pm0.2$ & $\bf{10/10}$ & $\bf{10/10}$ & $0.3\pm0.2$ & $0.3\pm0.2$ & $0.3\pm0.2$ & $\bf{10/10}$ & $\bf{10/10}$\\
\hline
\multicolumn{1}{c|}{\multirow{2}{*}{Small rectangular}}
& $1.0~\mathrm{\,cm}$ & $0.3\pm0.1$ & $0.1\pm0.1$ & $0.2\pm0.2$ & $\bf{10/10}$ & $\bf{10/10}$ & - & - & - & $ 6/10 $ & $ 6/10 $\\
& $2.5~\mathrm{\,cm}$ & $0.3\pm0.2$ & $0.1\pm0.1$ & $0.2\pm0.2$ & $\bf{10/10}$ & $\bf{10/10}$ & - & - & - & $ 0/10 $ & $ 0/10 $\\
            \hline
            \end{tabular}
    \endgroup
\vspace{-3.0mm}
\end{table*}

\begin{table*}[t]
    \centering
    \caption{
            The accuracy of stable placing.
            }
    \label{tab:result_stable_placement}
    \begingroup
    \scalefont{0.90}
        \begin{tabular}{c|c||cccm{6mm}m{7mm}|cccm{6mm}m{7mm}}
        \hline
            \multicolumn{1}{c|}{\multirow{2}{*}{\shortstack{Object \\ type}}} &
            \multicolumn{1}{c||}{\multirow{2}{*}{Object}} &
            \multicolumn{5}{c}{Tactile-based (Ours)} &
            \multicolumn{5}{|c}{F/T-based (Baseline)} \\
            \cline{3-12} 
            \multicolumn{1}{c|}{} & \multicolumn{1}{c||}{}
                             & \multirow{1}{*}{Roll [deg]}
                             & \multirow{1}{*}{Pitch [deg]}
                             & \multirow{1}{*}{All [deg]}
                             & \multirow{1}{*}{$< 1$ deg}
                             & \multirow{1}{*}{$< 2$ deg}
                             & \multirow{1}{*}{Roll [deg]}
                             & \multirow{1}{*}{Pitch [deg]}
                             & \multirow{1}{*}{All [deg]}
                             & \multirow{1}{*}{$< 1$ deg}
                             & \multirow{1}{*}{$< 2$ deg} \\
            \hline \hline
\multicolumn{1}{c|}{\multirow{6}{*}{\shortstack{Elongated \\ shape}}}
&Small rectangular & $0.4\pm0.2$ & $0.1\pm0.1$ & $0.2\pm0.2$ & $\bf{10/10}$ & $\bf{10/10}$ & - & - & - & $ 2/10 $ & $ 2/10 $\\
&Large rectangular & $0.4\pm0.3$ & $0.3\pm0.2$ & $0.4\pm0.3$ & $\bf{10/10}$ & $\bf{10/10}$ & $0.4\pm0.2$ & $0.4\pm0.3$ & $0.4\pm0.2$ & $\bf{10/10}$ & $\bf{10/10}$\\
&Wooden rectangular & $0.4\pm0.2$ & $0.2\pm0.1$ & $0.3\pm0.2$ & $\bf{10/10}$ & $\bf{10/10}$ & - & - & - & $ 7/10 $ & $ 7/10 $\\
&Wooden cylinder & $0.2\pm0.2$ & $0.1\pm0.1$ & $0.2\pm0.1$ & $\bf{10/10}$ & $\bf{10/10}$ & - & - & - & $ 6/10 $ & $ 7/10 $\\
&Beaker & $0.2\pm0.2$ & $0.3\pm0.1$ & $0.3\pm0.2$ & $\bf{10/10}$ & $\bf{10/10}$ & - & - & - & $ 4/10 $ & $ 4/10 $\\
&Metal rectangular & $0.2\pm0.1$ & $0.2\pm0.1$ & $0.2\pm0.1$ & $\bf{10/10}$ & $\bf{10/10}$ & $0.3\pm0.2$ & $0.1\pm0.1$ & $0.2\pm0.2$ & $\bf{10/10}$ & $\bf{10/10}$\\
\hline

\multicolumn{1}{c|}{\multirow{2}{*}{V shape}}
&V block & $0.4\pm0.3$ & $0.2\pm0.1$ & $0.3\pm0.2$ & $\bf{10/10}$ & $\bf{10/10}$ & - & - & - & $ 1/10 $ & $ 1/10 $\\
&Joint & $0.7\pm0.5$ & $0.2\pm0.1$ & $0.4\pm0.5$ & $\bf{8/10}$ & $\bf{10/10}$ & - & - & - & $ 0/10 $ & $ 0/10 $\\
\hline

\multicolumn{1}{c|}{\multirow{2}{*}{\shortstack{Asymmetry}}}
&LEGO blocks & $0.3\pm0.2$ & $0.9\pm0.1$ & $0.6\pm0.3$ & $\bf{10/10}$ & $\bf{10/10}$ & - & - & - & $ 6/10 $ & $ 6/10 $\\
&Metal object & $0.2\pm0.2$ & $0.2\pm0.1$ & $0.2\pm0.2$ & $\bf{10/10}$ & $\bf{10/10}$ & - & - & - & $ 0/10 $ & $ 0/10 $\\
\hline

\multicolumn{1}{c|}{\multirow{2}{*}{\shortstack{Non-surface\\contact}}}
&Metal pedestal & $0.2\pm0.2$ & $0.2\pm0.1$ & $0.2\pm0.1$ & $\bf{10/10}$ & $\bf{10/10}$ & $0.2\pm0.2$ & $0.1\pm0.1$ & $0.2\pm0.1$ & $\bf{10/10}$ & $\bf{10/10}$\\
&Conversion connector & $0.4\pm0.2$ & $0.2\pm0.1$ & $0.3\pm0.2$ & $\bf{10/10}$ & $\bf{10/10}$ & - & - & - & $ 2/10 $ & $ 2/10 $\\
\hline

\multicolumn{1}{c|}{\multirow{2}{*}{\shortstack{Soft \\ material}}}
&Wood bond tube & $0.3\pm0.2$ & $0.2\pm0.1$ & $0.2\pm0.2$ & $\bf{10/10}$ & $\bf{10/10}$ & - & - & - & $ 1/10 $ & $ 1/10 $\\
&Hand cream tube & $0.4\pm0.3$ & $0.4\pm0.2$ & $0.4\pm0.2$ & $\bf{10/10}$ & $\bf{10/10}$ & - & - & - & $ 4/10 $ & $ 4/10 $\\
\hline

\multicolumn{1}{c|}{\multirow{2}{*}{\shortstack{Fewer \\ contact}}}
&Test tube stand & $0.1\pm0.1$ & $0.6\pm0.2$ & $0.3\pm0.3$ & $\bf{10/10}$ & $\bf{10/10}$ & $0.2\pm0.2$ & $0.5\pm0.2$ & $0.3\pm0.2$ & $\bf{10/10}$ & $\bf{10/10}$\\
&Metal part & $0.4\pm0.2$ & $0.1\pm0.1$ & $0.3\pm0.2$ & $\bf{10/10}$ & $\bf{10/10}$ & $0.3\pm0.2$ & $0.1\pm0.1$ & $0.2\pm0.2$ & $\bf{10/10}$ & $\bf{10/10}$\\
\hline

Round shape
&Round bottom flask & $0.4\pm0.2$ & $0.2\pm0.2$ & $0.3\pm0.2$ & $\bf{10/10}$ & $\bf{10/10}$ & - & - & - & $ 0/10 $ & $ 0/10 $\\
\hline

Liquid
&Bottle with liquid & $0.4\pm0.3$ & $0.2\pm0.2$ & $0.3\pm0.2$ & $\bf{10/10}$ & $\bf{10/10}$ & - & - & - & $ 4/10 $ & $ 5/10 $\\

\hline
            \end{tabular}
    \endgroup
\vspace{-6.0mm}
\end{table*}
\subsection{Different Grasp Positions and Support Polygon Size}
\label{sec:Ablation Study}
We evaluate the influence of the object's grasping position and the size of the object's support polygon on stable placing.
To investigate these effects, we utilize two rectangular objects, small rectangular and large rectangular, which have the same height and different sizes of support polygon (Fig.~\ref{fig:target_objects}).
The dimensions of small and large rectangular are, respectively,
$30~\mathrm{\,mm} \times 30~\mathrm{\,mm} \times 70~\mathrm{\,mm}$, 
$50~\mathrm{\,mm} \times 50~\mathrm{\,mm} \times 70~\mathrm{\,mm}$ in terms of width, depth, and height.
These two objects were made by 3D printing, and they both weigh $142 \mathrm{\,g}$.
The two grasping positions are at heights of $1.0~\mathrm{\,cm}$ and $2.5~\mathrm{\,cm}$ from the desk.
To compare under the same noise conditions, the stable placing experiment, following the process outlined in \cref{sec:Process of object placing}, was conducted 10 times with the object tilted 10 degrees to the roll and pitch in the tactile-sensor coordinate in the pre-placing process (Fig.~\ref{fig:experiment_process}~(4)).
Table~\ref{tab:ablation_study} presents the mean and standard deviation of the roll and pitch misalignment and their combined values from ten iterations of stable placing experiments.
However, when the error magnitude exceeds 2 degrees, we do not compute the mean and standard deviation as the means appear exceedingly large when the object topples over.
We also report the number of times when roll and pitch errors were below 1 and 2 degrees.
We chose the thresholds of 1 degree and 2 degrees based on our experimental considerations. Specifically, the 1-degree threshold was selected due to the detection accuracy of the arUco marker.
The 2-degree threshold was chosen because our experimental experience suggested that exceeding this value often leads to the object toppling over.

Using our proposed tactile-based method, stable placing was achieved under all conditions regardless of the grasping position and the size of the support polygon.
In the case of the baseline F/T-based method, for the large rectangular object, stable placing was achieved at both grasp heights.
For the small rectangular object, while there were six successful outcomes with a low grasping height ($1.0~\mathrm{\,cm}$), there were no successes when the grasping height was high ($2.5~\mathrm{\,cm}$).

The reasons for this are explained below.
The direction of corrective rotation can be estimated through the sign of the torque. 
For a rectangular object, as illustrated in Fig.~\ref{fig:rotation_direction}~(2), bias in estimating the direction of corrective rotation depends on the grasping position and the size of the supporting polygon.
A low grasping position or large support polygon may still enable accurate estimation despite sensor inaccuracies since the area in which stable placement can be achieved is larger.
Conversely, a high grasping position or small support polygon decreases the area in which stable placing can be feasible and increases the risk of sensor inaccuracies negatively affecting the direction of corrective rotation estimation.
F/T sensors' measured values have low accuracy due to their noisy nature.
We note that a low pass filter is used to remove noise from the F/T sensor, but noise that cannot be removed still has an effect.
Therefore, even under the same conditions of conducting 10 stable placings with a 10-degree tilt in roll and pitch, failures may occasionally occur due to the inherent uncertainty of the F/T sensor.
Additionally, cable tension can cause a shift in the torque value, which could potentially lead to misinterpretations of the object's direction of corrective rotation.

\begin{figure*}[t]
    \centering
    \includegraphics[width=1.95\columnwidth]{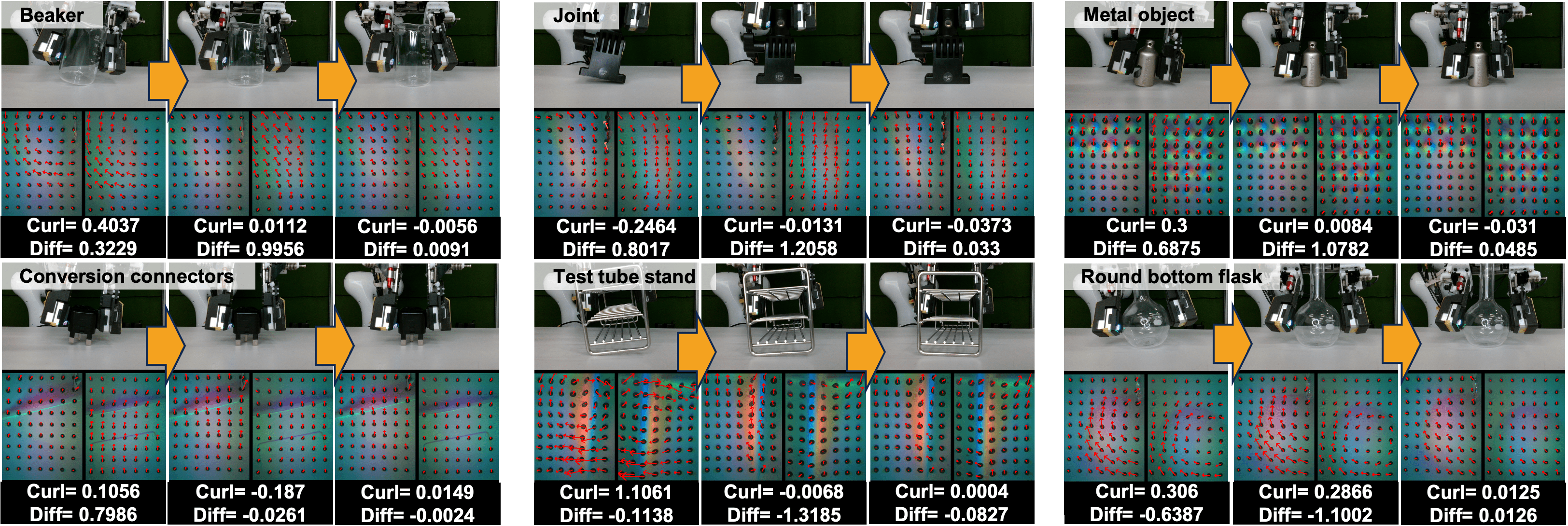}
    \caption{Examples of stable placing. For the sake of visibility, the length of the arrows is eight times the actual displacement.}
    \label{fig:stable_placing}
    \vspace{-7.0mm}
\end{figure*}
\subsection{Evaluation across Diverse Objects}
\label{sec:Comparison in Various Objects}
We evaluate the proposed method and a baseline F/T-based method for the stable placing of 18 different objects.
This stable placing experiment, which follows the process outlined in \cref{sec:Process of object placing}, was conducted 10 times with the object tilted.
The tilt in the pre-placing process (Fig.~\ref{fig:experiment_process}~(4)) is calculated from a normal distribution with a mean of $\pm10$ degrees (and the direction of the tilt is uniformly sampled to be positive or negative) and a variance of 1 in both roll and pitch directions in the tactile-sensor coordinate.
For objects prone to toppling over, we specify their tilts as follows:
The conversion connector, which is tilted even in the stable placing state, has a normally distributed noise with a mean of $\pm10$ degrees in the roll direction and a mean of $\pm3$ degrees in the pitch direction, while the Joint object has a mean of -5 degrees for noise in the pitch direction and a mean of 10 degrees in the roll direction.
For hand cream, which has a small support polygon, a mean of $\pm5$ degrees.
The variance of tilt for all items tested is 1.
The Table~\ref{tab:result_stable_placement} shows the results.

In the Tactile-based method, stable placing with an error of less than 1 degree was achieved for all objects except for the Joint object.
The method was successful for objects with asymmetrical shapes and textures, objects with a point of contact with the desk instead of a surface, soft objects, objects with a small size of support polygon, and objects with a changing center of gravity, such as liquids.
Fig.~\ref{fig:stable_placing} shows examples of the stable placing process using the proposed method.
The \textit{Curl} and \textit{Diff} values are nearly zero for a variety of objects, and stable placement is achieved.

Additionally, the method was also robust against cracks in the sensor of the surface and manufacturing differences in the size and shape of the black dots.
The image of the tactile sensor in Fig.~\ref{fig:target_objects} shows that the surface of the tactile sensor is cracked.
The tactile sensor was completely damaged and was replaced by another GelSight, which was also handled with a large black dot due to individual differences in the GelSight (Metal object and Test tube stand in Fig.~\ref{fig:stable_placing}).

For the Joint object, while it did not topple over, an error of less than 2 degrees remained.
We analyze this occurrence as follows:
The displacement of the black dots on GelSight is calculated based on their positions before stable placing when noise is added to the object's pose.
In the Joint object experiment, the object was grasped at the tip of the V-shape, which was far from the center of gravity.
As gravity causes the object to rotate, the black dots on GelSight deviate from their positions in the stable state.
Even when the Joint object reaches a stable placement, some displacement of the black point remains present.
Thus, the robot arm is controlled to apply force to the \textit{force application point} to reduce the displacement of the black point.
Unlike objects with a larger support polygon that retain their pose, the Joint object, with its smaller support polygon, appears to tilt due to the force intended for its rotation.
However, the tilt is small, less than 2 degrees, and the risk of toppling over is small.

The Tactile-based method achieved high accuracy for a variety of objects, while the F/T-based method often failed.
A common feature of the objects with a 100\% success rate in the F/T-based method is that the support polygon is large, such as large rectangular, metal rectangular, metal pedestal, test tube stand, and metal part.
Objects with small support polygon failed with probability or did not succeed at all.
As explained in \Cref{sec:Ablation Study}, the F/T sensor is affected by sensor noise and the cable tension, causing uncertainty in sensor values and misrecognizing the direction of rotation of the object.
As a result, there were many failures that resulted in the object toppling over in the F/T sensor-based approach.
There were also failures that the object's tilt was kept after contact between the object and the desk.
Moreover, some were seen where the robot arm applied force in the direction that the object was to be upright, but the object passed through the upright state and toppled over.

In summary, the F/T sensor method fails in stable placing due to sensor noise and the cable tension, causing deviations in the measured torque values, which result in misrecognition of the object's direction of corrective rotation.
On the other hand, the Tactile-based method succeeds in stable placing with high accuracy as a result of its precise estimation of the object's direction of corrective rotation.

\section{Conclusion} 
\label{sec:conclusion}
We propose a method for stable object placing using tactile sensors.
Under unstable object placing conditions, we observed that the displacement pattern of the black dots in GelSight aligns with the two possible directions of corrective rotation.
By estimating these two object directions of corrective rotation from \textit{Curl} and \textit{Diff}, the robot can manipulate the object's pose for stable placing.
This method of estimating the direction of corrective rotation is robust against sensor failures such as cracks in the surface, and manufacturing differences in the size and shape of the black dots.
In experiments with 18 objects with different characteristics, the F/T sensor-based method fails because it misrecognizes the direction of rotation for objects other than those with a large support polygon due to sensor noise and cable tension.
In contrast, the proposed Tactile sensor-based approach can achieve stable placing with high accuracy (less than 1-degree error) in nearly 100\% of cases across all objects, highlighting its potential as an effective F/T sensor alternative.
\section*{ACKNOWLEDGMENT}\small
The authors thank Avinash Ummadisingu and Dr. Naoki Fukaya for the many discussions about this research.
This work was supported by JST [Moonshot R\&D][Grant Number JPMJMS2033].
\bibliographystyle{IEEEtran} 
\bibliography{IEEEabrv,bibliography}

\begin{thebibliography}{10}
\providecommand{\url}[1]{#1}
\csname url@rmstyle\endcsname
\providecommand{\newblock}{\relax}
\providecommand{\bibinfo}[2]{#2}
\providecommand\BIBentrySTDinterwordspacing{\spaceskip=0pt\relax}
\providecommand\BIBentryALTinterwordstretchfactor{4}
\providecommand\BIBentryALTinterwordspacing{\spaceskip=\fontdimen2\font plus
\BIBentryALTinterwordstretchfactor\fontdimen3\font minus \fontdimen4\font\relax}
\providecommand\BIBforeignlanguage[2]{{%
\expandafter\ifx\csname l@#1\endcsname\relax
\typeout{** WARNING: IEEEtran.bst: No hyphenation pattern has been}%
\typeout{** loaded for the language `#1'. Using the pattern for}%
\typeout{** the default language instead.}%
\else
\language=\csname l@#1\endcsname
\fi
#2}}

\bibitem{raibert1981hybrid}
M.~H. Raibert and J.~J. Craig, ``{Hybrid position/force control of manipulators},'' 1981.

\bibitem{yuan2017gelsight}
W.~Yuan, \emph{et~al.}, ``{GelSight: High-Resolution Robot Tactile Sensors for Estimating Geometry and Force},'' \emph{Sensors}, vol.~17, no.~12, p. 2762, 2017.

\bibitem{anzai2020deep}
T.~Anzai and K.~Takahashi, ``{Deep gated multi-modal learning: In-hand object pose changes estimation using tactile and image data},'' in \emph{2020 IEEE/RSJ International Conference on Intelligent Robots and Systems (IROS)}.\hskip 1em plus 0.5em minus 0.4em\relax IEEE, 2020, pp. 9361--9368.

\bibitem{bauza2023tac2pose}
M.~Bauza, \emph{et~al.}, ``{Tac2pose: Tactile object pose estimation from the first touch},'' \emph{The International Journal of Robotics Research}, vol.~42, no.~13, pp. 1185--1209, 2023.

\bibitem{dong2021tactile}
S.~Dong, \emph{et~al.}, ``{Tactile-rl for insertion: Generalization to objects of unknown geometry},'' in \emph{2021 IEEE International Conference on Robotics and Automation (ICRA)}.\hskip 1em plus 0.5em minus 0.4em\relax IEEE, 2021, pp. 6437--6443.

\bibitem{zhang2019effective}
Y.~Zhang, \emph{et~al.}, ``{Effective estimation of contact force and torque for vision-based tactile sensors with helmholtz--hodge decomposition},'' \emph{IEEE Robotics and Automation Letters}, vol.~4, no.~4, pp. 4094--4101, 2019.

\bibitem{sun2022onepose}
J.~Sun, \emph{et~al.}, ``{Onepose: One-shot object pose estimation without cad models},'' in \emph{Proceedings of the IEEE/CVF Conference on Computer Vision and Pattern Recognition}, 2022, pp. 6825--6834.

\bibitem{Li2023vox}
B.~Li, \emph{et~al.}, ``{VoxDet: Voxel Learning for Novel Instance Detection},'' in \emph{Proceedings of the Advances in Neural Information Processing Systems (NeurIPS)}, 2023.

\bibitem{chavan2018stable}
N.~Chavan-Dafle and A.~Rodriguez, ``Stable prehensile pushing: In-hand manipulation with alternating sticking contacts,'' in \emph{2018 IEEE International Conference on Robotics and Automation (ICRA)}.\hskip 1em plus 0.5em minus 0.4em\relax IEEE, 2018, pp. 254--261.

\bibitem{von2020contact}
F.~Von~Drigalski, \emph{et~al.}, ``{Contact-based in-hand pose estimation using bayesian state estimation and particle filtering},'' in \emph{2020 IEEE International Conference on Robotics and Automation (ICRA)}.\hskip 1em plus 0.5em minus 0.4em\relax IEEE, 2020, pp. 7294--7299.

\bibitem{pankert2023learning}
J.~Pankert and M.~Hutter, ``{Learning Contact-Based State Estimation for Assembly Tasks},'' in \emph{2023 IEEE/RSJ International Conference on Intelligent Robots and Systems (IROS)}.\hskip 1em plus 0.5em minus 0.4em\relax IEEE, 2023, pp. 5087--5094.

\bibitem{lach2023placing}
L.~Lach, \emph{et~al.}, ``{Placing by Touching: An empirical study on the importance of tactile sensing for precise object placing},'' in \emph{2023 IEEE/RSJ International Conference on Intelligent Robots and Systems (IROS)}.\hskip 1em plus 0.5em minus 0.4em\relax IEEE, 2023, pp. 8964--8971.

\bibitem{pai2023laboratory}
S.~Pai, \emph{et~al.}, ``{Laboratory Automation: Precision Insertion with Adaptive Fingers utilizing Contact through Sliding with Tactile-based Pose Estimation},'' \emph{arXiv preprint arXiv:2309.16170}, 2023.

\bibitem{ota2023tactile}
K.~Ota, \emph{et~al.}, ``{Tactile Estimation of Extrinsic Contact Patch for Stable Placement},'' \emph{arXiv preprint arXiv:2309.14552}, 2023.

\bibitem{whitney1982quasi}
D.~E. Whitney \emph{et~al.}, ``{Quasi-static assembly of compliantly supported rigid parts},'' \emph{Journal of Dynamic Systems, Measurement, and Control}, vol. 104, no.~1, pp. 65--77, 1982.

\bibitem{higuera2023neural}
C.~Higuera, \emph{et~al.}, ``{Neural contact fields: Tracking extrinsic contact with tactile sensing},'' in \emph{2023 IEEE International Conference on Robotics and Automation (ICRA)}.\hskip 1em plus 0.5em minus 0.4em\relax IEEE, 2023, pp. 12\,576--12\,582.

\bibitem{shimizu2002spatial}
M.~Shimizu and K.~Kosuge, ``Spatial parts mating with fiction using structured compliance with compliance center,'' in \emph{IEEE/RSJ International Conference on Intelligent Robots and Systems}, vol.~2.\hskip 1em plus 0.5em minus 0.4em\relax IEEE, 2002, pp. 1585--1590.

\bibitem{gelsight2024}
``Gelsight mini,'' \url{https://www.gelsight.com/gelsightmini/}, 2024.

\bibitem{garrido2014automatic}
S.~Garrido-Jurado, \emph{et~al.}, ``Automatic generation and detection of highly reliable fiducial markers under occlusion,'' \emph{Pattern Recognition}, vol.~47, no.~6, pp. 2280--2292, 2014.

\end{thebibliography}
\end{document}